\documentclass[review]{elsarticle}
\graphicspath{ {./figures/} }
\usepackage{float}
\usepackage{verbatim}
\usepackage{apalike}
\restylefloat{figure}
\restylefloat{table}

\usepackage{lineno}
\usepackage{mathtools}
\usepackage{rotating}
\usepackage{xcolor}
 \usepackage{url}
\usepackage[
colorlinks=true,
urlcolor=blue,
linkcolor=blue,
citecolor=blue
]{hyperref}
\usepackage{etoolbox}
\usepackage{ragged2e}
\usepackage{booktabs}
\usepackage{graphicx}
\graphicspath{{figures/}}
\usepackage{threeparttable}
\usepackage{multirow}
\usepackage{booktabs}
\usepackage{siunitx}
\usepackage{bm,amsmath}
\usepackage{amssymb}
\usepackage{caption}
\usepackage{subcaption}
\usepackage{soul}
\usepackage{enumitem}

\journal{arXiv}

\biboptions{authoryear}

\begin{document}
\begin{frontmatter}

\begin{titlepage}
\begin{center}
\vspace*{1cm}

\textbf{Learning ECG Signal Features Without Backpropagation \\ Using Linear Laws}

\vspace{1.5cm}

P\'eter P\'osfay$^{a}$ (posfay.peter@wigner.hun-ren.hu, ORCID: 0000-0002-6769-3302), Marcell T.\ Kurbucz$^{a,b}$ (kurbucz.marcell@wigner.hun-ren.hu, ORCID: 0000-0002-0121-6781), P\'eter Kov\'acs$^{c}$ (kovika@inf.elte.hu, ORCID: 0000-0002-0772-9721), Antal Jakov\'ac$^{a,b}$ (jakovac.antal@wigner.hun-ren.hu, ORCID: 0000-0002-7410-0093) \\

\hspace{10pt}

\begin{flushleft}
\small  
$^a$ Department of Computational Sciences, Institute for Particle and Nuclear Physics, HUN-REN Wigner Research Centre for Physics, 29-33 Konkoly-Thege Mikl\'os Street, H-1121 Budapest, Hungary \\
$^b$ Department of Statistics, Institute of Data Analytics and Information Systems, Corvinus University of Budapest, 8 F\H{o}v\'am Square, H-1093 Budapest, Hungary \\
$^c$ Department of Numerical Analysis, E\"otv\"os Lor\'and University, 1/c. P\'azm\'any P\'eter s\'et\'any, H-1117 Budapest, Hungary

\vspace{1cm}
\textbf{Corresponding Author:} \\
Marcell T.\ Kurbucz \\
Department of Computational Sciences, Institute for Particle and Nuclear Physics, HUN-REN Wigner Research Centre for Physics, 29-33 Konkoly-Thege Mikl\'os Street, H-1121 Budapest, Hungary; Department of Statistics, Institute of Data Analytics and Information Systems, Corvinus University of Budapest, 8 F\H{o}v\'am Square, H-1093 Budapest, Hungary \\
Tel: (+36) 1 392 2222 \\
ORCID: 0000-0002-0121-6781\\
Email: kurbucz.marcell@wigner.hun-ren.hu

\end{flushleft}        
\end{center}
\end{titlepage}

\title{Learning ECG Signal Features Without Backpropagation Using Linear Laws}

\author[label1]{P\'eter P\'osfay}
\ead{posfay.peter@wigner.hun-ren.hu}

\author[label1,label2]{Marcell T. Kurbucz\corref{cor1}}
\ead{kurbucz.marcell@wigner.hun-ren.hu}

\author[label3]{P\'eter Kov\'acs}
\ead{kovika@inf.elte.hu}

\author[label1]{Antal Jakov\'ac}
\ead{jakovac.antal@wigner.hun-ren.hu}

\cortext[cor1]{Corresponding author.}
\address[label1]{Department of Computational Sciences, Institute for Particle and Nuclear Physics, HUN-REN Wigner Research Centre for Physics, 29-33 Konkoly-Thege Mikl\'os Street, H-1121 Budapest, Hungary}
\address[label2]{Institute of Data Analytics and Information Systems, Corvinus University of Budapest, 8 F\H{o}v\'am Square, H-1093 Budapest, Hungary}
\address[label3]{Department of Numerical Analysis, E\"otv\"os Lor\'and University, 1/c P\'azm\'any P\'eter s\'et\'any, H-1117 Budapest, Hungary}

\begin{abstract}
This paper introduces LLT-ECG, a novel method for electrocardiogram (ECG) signal classification that leverages concepts from theoretical physics to automatically generate features from time series data. Unlike traditional deep learning approaches, LLT-ECG operates in a forward manner, eliminating the need for backpropagation and hyperparameter tuning. By identifying linear laws that capture shared patterns within specific classes, the proposed method constructs a compact and verifiable representation, enhancing the effectiveness of downstream classifiers. We demonstrate LLT-ECG's state-of-the-art performance on real-world ECG datasets from PhysioNet, underscoring its potential for medical applications where speed and verifiability are crucial.
\end{abstract}


\end{frontmatter}

\section{Introduction}
\label{sec}

\noindent
The success of machine learning systems hinges heavily on how data are represented and processed before being fed into models. The optimal data representation is often task-specific and may evolve as the problem or data changes. A significant portion of machine learning research has been devoted to improving data representations through feature engineering---manually transforming raw data into formats better suited for learning algorithms. While these handcrafted techniques can boost model performance, they have inherent limitations. Feature engineering often relies on human intuition, which can lead to suboptimal or non-generalizable representations. As the scale and complexity of data increase, these methods also become labor-intensive and less effective, highlighting the need for more automated approaches to data preprocessing and feature extraction \citep{RepL_review}.

To address the challenges posed by increasing data complexity and variety, modern machine learning has shifted towards methods that enable algorithms to learn data representations autonomously. This paradigm shift has yielded impressive advances across various fields, including speech recognition \citep{replearn_speech1, replearn_speech2} and image recognition \citep{replearn_success5}, where models are now capable of dynamically learning hierarchical representations from raw data. Hierarchical data organization has also led to breakthroughs in natural language processing \citep{replearn_success6, replearn_success7}, demonstrating its versatility across domains. Additionally, this concept has roots in neuroscience, particularly in visual information processing \citep{visneuroscience}, where hierarchies play a crucial role in the brain's ability to interpret complex visual stimuli. This idea has inspired the design of convolutional neural networks (CNNs), which mimic the hierarchical structure of the human visual cortex \citep{cnn_lecun}. CNNs have become a cornerstone of computer vision, demonstrating the power of data-driven representation learning.

Despite the success of neural networks, including CNNs, they are not without drawbacks. Neural networks can suffer from issues such as overfitting, the vanishing gradient problem, and resource-intensiveness. Furthermore, their performance often hinges on large amounts of labeled data and extensive computational resources, making them difficult to deploy in settings with limited data or processing power. Additionally, neural networks can lack transparency, creating challenges in applications where this is essential, such as in healthcare \citep{kemker2018measuring, hanin2018neural}.

The proposed LLT-ECG method overcomes the challenges of feature extraction by learning meaningful features from raw ECG data without manual intervention or extensive hyperparameter tuning.\footnote{The LLT method is inspired by the principles of representation learning and theoretical physics, specifically drawing on the concept of renormalization in physics, where complex systems are simplified by disentangling underlying factors \citep{renorm1, renorm2, renorm3}.} LLT-ECG identifies linear relationships, or 'linear laws,' that capture shared patterns within specific classes, creating a compact latent representation of the input data. The proposed feature-learning procedure is mathematically transparent, which is particularly valuable in medical applications where understanding the basis of a model’s predictions is essential (\cite{Lipton}). Moreover, unlike CNNs, this approach does not rely on backpropagation or large amounts of labeled data, making it especially suitable for medical applications where verifiability, speed, and resource efficiency are crucial~\citep{ECG_data_analysis_paper1}.

While our prior research on the LLT method focused on general time series classification tasks \citep{LLT, kurbucz2024llt}, this paper specifically adapts and extends LLT for the classification of ECG signals, which present unique challenges not addressed in our previous work. In earlier applications, LLT performed transformations on complete, raw time series data without specialized preprocessing steps. However, ECG signals require tailored preprocessing due to their periodic nature and the prominence of spike-like features such as QRS complexes.

\vspace{1em}

\noindent
The main contributions of this paper are as follows:
\vspace{-1em}
\begin{sloppypar}
\begin{itemize}[noitemsep] 
\item We propose a novel ECG classification method, LLT-ECG, which integrates ECG signal preprocessing with a semi-supervised feature extraction technique inspired by the conservation laws of physics.
\item We evaluate the effectiveness of LLT-ECG by comparing it to a wide range of methods on real-world ECG datasets from PhysioNet \citep{goldberger2000physiobank}.
\item Our proposed method achieves state-of-the-art performance with a hyperparameter-free, lightweight design, eliminating the need for backpropagation.
\item We demonstrate that the LLT transformation of ECG signals typically produces feature spaces that are nearly linearly separable, enabling high classification accuracy even with basic linear classifiers.
\end{itemize}
\end{sloppypar}

The rest of this paper is structured as follows. Section~\ref{sec:ecgclass} provides a review of existing work on the ECG classification problem, along with a high-level overview of the proposed method in the context of the current literature. In Section~\ref{sec:math_int}, we describe the mathematical foundations of linear laws and explain the feature extraction process. Section~\ref{sec:featurelearn} discusses how linear laws are able to transform datasets and extract class-specific features for multiclass classification and anomaly detection. Section~\ref{sec:data} details the datasets used for evaluation. Finally, results are presented and discussed in Section~\ref{sec:results}  and \ref{sec:disc}, respectively, followed by conclusions and future research directions in Section~\ref{sec:conclusion}.

\section{ECG Classification} 
\label{sec:ecgclass}

\subsection{Related Work}

\noindent
Computer-assisted ECG analysis has a long history \citep{survey} and remains an active area of research, driven by the increasing volume of data generated by modern technologies. Advances in data acquisition tools---such as the Internet of things (IoT), wearable devices, and smart technologies \citep{ECG_sensors}---have facilitated commercial applications in ECG monitoring. These innovations have resulted in the generation of vast datasets that necessitate efficient storage, processing, and interpretation. Key challenges within this field include data labeling for training purposes and the extraction of pertinent features that encapsulate medical information while mitigating noise interference. Data labeling is typically labor-intensive and costly, thereby constraining the availability of training datasets. To address this limitation, self-supervised learning methodologies have been developed \citep{selfsup}. Concurrently, feature extraction remains a complex endeavor, achievable either through manual feature engineering or via automated representation learning techniques \citep{kiranyaz, Kovacs_data}.

\subsubsection{Feature Extraction in Clinical Settings}

\noindent
In clinical settings, decision-making processes must be transparent, often leading to a preference for handcrafted features that combine medical expertise with signal processing knowledge. Temporal and statistical features---such as RR intervals, ECG durations, and moment-based indices---are widely studied in the literature \citep{opt_ecg_feat_selection, survey}. Additionally, morphological features related to the shapes of key ECG waveforms (including the QRS complex, T-wave, and P-wave) are critical, especially for detecting arrhythmias, which is the primary focus of this section \citep{ECG_data_analysis_book2}.

\subsubsection{Techniques for Morphological Feature Extraction}

\noindent
Several methodologies exist for extracting morphological features from ECG signals. Time-domain approaches operate directly on ECG signal samples \citep{subsample_features}, calculating metrics such as power, derivatives, and extreme values \citep{JEKOVA2008248}. However, these features are often susceptible to noise, prompting a frequent transition to frequency-domain feature extraction. Spectral methods, such as linear filtering, presume weak stationarity of the signal---a condition frequently unmet in ECG data due to physiological variabilities like respiration, body movement, and arrhythmia. To accommodate the nonstationary nature of ECG signals, joint time-frequency representations have been proposed, including the short-time Fourier transform, Choi–Williams distribution, and multiwindow spectrogram \citep{Cakrak2001}. Unlike these techniques, wavelet transformations employ time windows of varying widths, thereby enhancing temporal and spectral localization of nonstationary features. This flexibility has rendered wavelets a preferred tool for ECG signal processing \citep{addison}.

\subsubsection{Adaptive Data-Driven Transformations}

\noindent
Despite the availability of diverse joint time-frequency representations, the foundational components---such as window functions or mother wavelets---are typically predetermined, which can be restrictive in ECG analysis where morphological features exhibit temporal and inter-individual variability. To overcome this limitation, adaptive data-driven transformations have been introduced. These approaches represent ECG signals using a series of basis functions optimized according to specific criteria. Techniques like variable projections with parameterized basis functions (e.g., Hermite, spline, and rational functions) offer optimal representations in the least-squares sense \citep{genvarpro, BOGNAR2020102034}. In contrast, principal component analysis (PCA) transforms data by identifying orthogonal directions that maximize variance \citep{ecgpca}, while independent component analysis (ICA) seeks to maximize higher-order statistics, such as kurtosis, to achieve blind source separation \citep{ica_vs_pca}. Linear transformations, including PCA and ICA, which map input features into well-separated spaces, can also be derived from data-driven learning. For instance, the neural dynamic classification algorithm \citep{NDC} constructs a feature map that minimizes intra-class variance while maximizing inter-class variance.

\subsection{Proposed Classification Pipeline}

\noindent
The proposed ECG classification method, LLT-ECG, consists of the following stages:

\begin{enumerate}[noitemsep]
\item \textbf{Preprocessing}: The dataset is segmented into various subsets based on lead differentiation and beat classification scenarios, followed by standard noise filtering and data normalization procedures.

\item \textbf{Learning Phase}: Linear laws are fitted to the normal class within the training set, capturing the underlying factors that characterize normal ECG patterns in the dataset.

\item \textbf{Feature Extraction}: The established linear laws are utilized to transform the data into a new representation, thereby enhancing features pertinent to classification. This step builds upon previous work involving linear laws \citep{LLT, LLT_mech} and introduces the LLT method, specifically tailored to accommodate the spike-like characteristics of ECG signals.

\item \textbf{Classification}: Classifiers are trained within the transformed feature space, and their performance is subsequently evaluated using the test set.
\end{enumerate}

This structured pipeline integrates the LLT algorithm with robust classifiers, enabling accurate and lightweight ECG analysis. Furthermore, this approach offers potential for developing transparent AI methods for ECG classification (see proposed future work in Section \ref{sec:conclusion}).

\section{Mathematical Background}
\label{sec:math_int}

\noindent
Here, we briefly summarize the mathematical foundations necessary for understanding linear laws \citep{LLT, LLT_mech, Jakovac_time_series, Jakovac_understanding} and implementing the LLT-type transformation applied to ECG signals.

Consider a time series \( y: \mathbb{R} \rightarrow \mathbb{V} \), where \( \mathbb{V} \) is a finite-dimensional Hilbert space. In practice, we work with finite-dimensional representations; thus, we assume that a faithful finite-dimensional representation of the time series is provided. Consequently, we can construct a finite set of \( n \)-length samples from this time series:

\begin{equation}
    \mathcal{Y} = \left\{ 
    Y^{(k)} \in \mathbb{V}^{n+1} \; \bigg| \;
    Y_{i}^{(k)} = y(t_{k} - i \Delta t), \; i \in \{0,\dots,n\}, \; k \in \{n,\dots,K\}
    \right\},
    \label{eq:Yki}
\end{equation}

\noindent
where \( K, n \in \mathbb{N} \), \( \Delta t \) is the sampling interval of the time series, and \( y(t_{k}) \) denotes the value of the time series at time \( t_{k} \). This construction is known as the time delay embedding of the time series, which is sufficient to capture the dynamic state of the system \citep{Takens1, Takens2}. The selection of \( t_{k} \) values that serve as base points for the \( n \)-length samples can be adapted to suit specific applications. In this work, we generate maximally overlapping \( n \)-length samples for a time series of length \( L \).

\subsection{Linear Laws}

\noindent
First, consider mappings of the following type:

\begin{equation}
    \mathcal{F}: \mathbb{V}^{n+1} \rightarrow \mathbb{R}, \quad \mathcal{F}\left(Y^{(k)}\right) = 0, \quad \forall \, k.
    \label{eq:F_mapping}
\end{equation}

\noindent
In this study, we focus exclusively on linear mappings. This assumption constrains the form of $\mathcal{F}$ in Eq.~\eqref{eq:F_mapping}. For convenience, we introduce matrix notation for the embedded time series $Y_{i}^{(k)}$ from Eq.~\eqref{eq:Yki}:

\begin{equation}
    Y_{ki} = Y^{(k)}_{i} = y(t_{k} - i \Delta t).
    \label{eq:Yki_matrix}
\end{equation}

Using this notation, the linear mapping $\mathcal{F}$ can be expressed as:

\begin{equation}
    \mathcal{F}(Y^{(k)}) = \sum_{i=0}^{n} Y_{ki} w_{i} \equiv (Y \mathbf{w})_{k} = 0, \quad \forall \, k,
    \label{eq:linear_mapping}
\end{equation}

\noindent
where $\mathbf{w}$ is a weight vector of length $n+1$. We refer to this construction as a ``linear law'' $\mathcal{F}$, represented by the vector $\mathbf{w}$. As mentioned in the introduction, the intuition behind this nomenclature originates from physics: $\mathcal{F}$ can be considered a ``law'' on the set $Y$ because it satisfies Eq.~\eqref{eq:linear_mapping}. This represents an ideal scenario that cannot be achieved empirically due to factors such as noise. Therefore, we utilize linear mappings that satisfy Eq.~\eqref{eq:linear_mapping} with a random quantity $\xi$ instead of zero, where $\langle \xi \rangle = 0$. We refer to these linear mappings as ``laws''.

\subsection{Determining Linear Laws from Data}

\noindent
To determine the coefficients \( w_{i} \) of the linear law, we express Eq.~\eqref{eq:linear_mapping} as \( \| Yw \| = 0 \). Using the standard quadratic norm, this becomes:

\begin{equation}
    \| Yw \|^2 = \frac{1}{K} (Yw)^T (Yw) = w^T C w = 0,
    \label{eq:pca_least_sq}
\end{equation}

\noindent
where

\begin{equation}
    C = \frac{1}{K} Y^T Y,
    \label{eq:corrmat}
\end{equation}

\noindent
is the correlation matrix of the dataset.

To avoid the trivial solution \( w = 0 \), we impose \( \| w \| = 1 \), transforming the problem into a constrained minimization task. Using Lagrange multipliers, we define:

\begin{equation}
    \chi^2(\lambda) = w^T C w - \lambda (w^T w - 1) \rightarrow \text{minimize},
    \label{eq:minimization}
\end{equation}

\noindent
where \( \lambda \) is the Lagrange multiplier. The solution satisfies the eigenvalue equation

\begin{equation}
    Cw^{(\lambda)} = \lambda w^{(\lambda)},
    \label{eq:pca_eigensystem}
\end{equation}

\noindent
yielding \( n+1 \) eigenvectors with corresponding eigenvalues \( \lambda \). To select the optimal eigenvector, consider that an exact linear law satisfies \( \| Yw \| = 0 \). However, due to noise, the laws yield a non-zero quantity:

\begin{equation}
    \mathcal{F}(Y^{(k)}) = \sum_{i=0}^n Y_{ki} w_i \equiv \xi_k,
    \label{eq:xi_def}
\end{equation}

\noindent
where \( \langle \xi \rangle = 0 \). Substituting into Eq.~\eqref{eq:pca_least_sq} gives:

\begin{equation}
    \| Yw \|^2 = \frac{1}{K} \sum_{k=1}^K \xi_k^2 = \left\langle \xi^2 \right\rangle.
    \label{eq:xi_sq}
\end{equation}

\noindent
Meanwhile, from Eq.~\eqref{eq:pca_eigensystem}, we have

\begin{equation}
    \| Yw \|^2 = \lambda w^T w = \lambda.
    \label{eq:xi_sq_2}
\end{equation}

Comparing Eqs.~\eqref{eq:xi_sq} and \eqref{eq:xi_sq_2}, the variance \( \left\langle \xi^2 \right\rangle \) equals \( \lambda \). Therefore, the optimal linear law corresponds to the eigenvector with the smallest eigenvalue, minimizing the variance and approximating \( \| Yw \| = 0 \). This eigenvector is guaranteed to exist as \( C \) is symmetric and positive definite.

The linear law is similar to the PCA method, but instead of the largest eigenvalue, we select the smallest. Thus, \( w_{i} \) is orthogonal to the dataset, indicating minimal variation in the direction of \( w \). This identifies a common property, analogous to the normal vector of a hyperplane fitting the data.

In practice, the laws are never exact due to variations such as noise in the data. Consequently, the values of the optimal laws fall within a narrow range around zero for each respective class. When used to generate features, this results in class elements clustering near zero, while non-class elements are positioned farther from zero. This property provides a natural framework for classification algorithms.

Our formulation assumes a long time series sample, embedded into the matrix $Y_{ki}$. In many cases, however, the learning set consists of labeled time series samples $y_m(t)$, where $m \in \mathbb{N}$. This situation can be reduced to the previous case. By applying the embedding process to each sample, we generate matrices $Y_{k'i'}^{(m)}$, which can be concatenated along their first axis (rows) to form a compound matrix $Y_{ki}$. Since the rows represent time-embedded samples and their order is irrelevant for calculating the linear law, this concatenation simply increases the learning set.

In standard notation, this augmentation forms a larger block matrix with extended rows. If each sample has a uniform length $L$ and the embedding depth is $I$, the matrix $Y_{k'i'}^{(m)}$ has dimensions $(L-I+1) \times I$. For $M$ samples, concatenating them yields a matrix $Y_{ki}$ with dimensions $((L-I+1)\cdot M) \times I$:
\begin{equation}
    Y_{ki} = 
    \begin{bmatrix}
        Y_{k'i'}^{(1)} \\
        Y_{k'i'}^{(2)} \\
        \vdots \\
        Y_{k'i'}^{(m)}
    \end{bmatrix}.
    \label{eq:augmented_Y}
\end{equation}
\noindent
This augmented matrix can be treated as $Y_{ki}$ in the previous formulation. Thus, the linear law determination process remains consistent, independent of the dataset's structure, as all samples are embedded into a unified matrix.

\section{Feature Learning via Linear Laws}
\label{sec:featurelearn}

\noindent
In the previous sections, we discussed how to determine linear laws given a set of samples. This section demonstrates how linear laws can be utilized to transform datasets and learn features from samples within a class, as well as how this technique can be applied to multiclass classification and anomaly detection problems.

We assume that linear laws encapsulate properties common to the elements of a defining set. This is formalized as a mapping from classes to linear laws:

\begin{equation}
    \mathcal{H}:  C_{j} \quad  \rightarrow  \quad w_{i}^{(j)},
    \label{eq:H_mapping}
\end{equation}

\noindent
where $C_{j}$ represents the sets corresponding to the classes in the dataset ($j \in \mathbb{N}$). The elements of a class can be transformed by linear laws as follows. Let $y_{m}(t)$ denote a sample from a given class, which is a time series. First, the time series is embedded in the standard manner, as described in Eqs.~\eqref{eq:Yki} and \eqref{eq:Yki_matrix}. This maps the $L$-length time series sample into an $(L-n+1) \times n$ matrix $Y_{ki}^{(m)}$. Then, the linear law $w_{i}^{(j)}$ is applied to the sample, analogous to Eq.~\eqref{eq:xi_def}:

\begin{equation}
    \sum_{i=0}^n Y_{ki}^{(m)} w_i^{(j)} =  \xi_k^{(m, j)}, 
    \label{eq:def_features}
\end{equation}

\noindent
where the vector $\xi_k^{(m, j)}$ contains the transformed features of sample $m$ according to the linear law $w_{i}^{(j)}$, which corresponds to class $C_{j}$. As illustrated by Eq.~\eqref{eq:xi_def}, these features quantify how effectively a linear law can transform the subsamples of a given sample to zero. The closer the transformed subsamples are to zero, the better the linear law describes the sample. Intuitively, the linear law $w_{i}^{(j)}$ transforms elements of class $C_{j}$ closer to zero than samples from other classes. Consequently, the features resulting from the transformation with $w_{i}^{(j)}$ act as similarity detectors for elements of class $C_{j}$.

\subsection{Multiclass Classification}

\noindent
The LLT features defined in Eq.~\eqref{eq:def_features} can be utilized for multiclass classification as follows. To assess a sample's similarity to each class, it must be transformed by the linear laws of all classes. Conceptually, when transforming an unknown sample, it is necessary to apply all linear laws corresponding to the possible classes to facilitate comparison. Using this transformed information, a classifier can be trained to recognize how elements of each class appear after being transformed by the linear laws of other classes. If the linear laws of different classes produce distinct transformations, the resulting feature vectors will be significantly different, especially when using the correct class's linear law. This separation in the abstract feature space enhances the performance of classifier algorithms by increasing the distance between samples from different classes.

Let the $j$-th class (and its corresponding linear law, as per Eq.~\eqref{eq:H_mapping}) be indexed by $j$. Then, the transformed features for a sample $y_{m}(t)$ can be organized into a feature vector as follows:

\begin{equation}
    \bm{\xi}^{m} = \left[\xi_k^{(m, 1)}, \xi_k^{(m, 2)}, \dots, \xi_k^{(m, J)} \right]. 
    \label{eq:transformed_features}
\end{equation}

\noindent
This vector comprises the feature vectors corresponding to each class.\footnote{The feature generation approach in the LLT-ECG method presented in this paper differs from the original LLT algorithm \citep{LLT}. In this method, we segment the signals to detect deviations from healthy patterns, whereas the original LLT analyzes the entire series using statistical measures such as mean and variance.} Depending on the application, this feature vector can be processed differently. One approach is to flatten $\bm{\xi}^{m}$ by concatenating the individual $\xi_k^{(m, j)}$ sequentially. Generally, this results in a feature vector of length $(L-n+1) \cdot J$, where $J$ is the number of classes and $n$ is the length of the linear law. In practice, this feature vector can be downsampled if a less detailed representation suffices for classification.

\subsection{Anomaly Detection}

\noindent
In binary classification scenarios ($J = 2$), the feature vector $\bm{\xi}$ can be significantly simplified. It suffices to use the linear law corresponding to the reference class:

\begin{equation}
    \bm{\xi}^{m} = \xi_k^{(m, 1)}. 
    \label{eq:transformed_features_simple}
\end{equation}

\noindent
This simplification is particularly advantageous for anomaly detection, where a well-defined reference class is provided, and the objective is to distinguish samples from this class from others that may not belong to any predefined class. By using only the reference class's linear law, the classifier focuses on determining whether a given sample resembles the reference class. This approach can also be considered a form of outlier detection, as it does not require a linear law for the outliers, which may not be well-defined or existent. Nonetheless, some outlier samples are necessary to train the classifier after the LLT transformation.

For example, in classifying ECG signals between normal and ectopic types, numerous ectopic variations exist, but the goal is not to classify these variations individually. Instead, the task is to identify whether an ECG signal is healthy. Normal heartbeats are selected as the reference class, and ectopic signals are classified as ``not normal.'' Each sample is transformed using the linear law derived from normal heartbeats, and the resulting features quantify the similarity of a sample to normal ECG signals. This constitutes a semi-supervised feature learning approach in which only a subset of the data, specifically the reference class samples, needs to be labeled.

\section{Employed Datasets}
\label{sec:data}

\noindent
In this study, we utilized two datasets: TwoLeadECG \citep{TwoLeadECG} and Variable Projection Networks (VPNet) \citep{Kovacs_data}. Both datasets are derived from the PhysioNet MIT-BIH Arrhythmia Database \citep{MIT_database}.

\subsection{TwoLeadECG Dataset}

\noindent
To demonstrate the transformation capabilities of the LLT algorithm in the ECG context---as applied by the proposed LLT-ECG approach---we employed the TwoLeadECG dataset, a subset of the MIT-BIH Long-Term ECG Database \citep{goldberger2000physiobank}. This dataset comprises long-term recordings from the same patient using two different leads. It contains a total of 1,162 samples, with 581 labeled as Class 1 and 581 labeled as Class 2. Each sample consists of 82 consecutive measurements. The classification task involves distinguishing between signals originating from each lead. Following the approach of \cite{harada2019biosignal}, we randomly divided the dataset into training and test sets, resulting in 523 signals for training and 639 signals for testing.

\subsection{VPNet Dataset}
\label{sec:vpnet_dataset}

\begin{figure}[ht]
     \centering
     \begin{subfigure}[b]{0.45\linewidth}
         \centering
         \includegraphics[width=\textwidth]{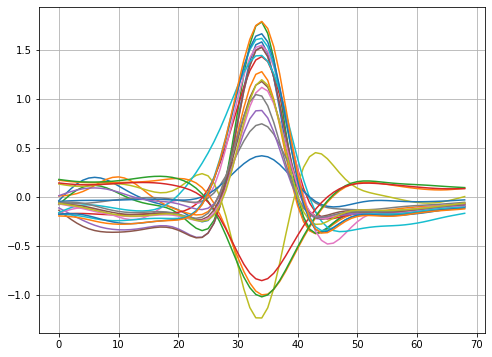}
         \caption{Healthy QRS Complexes}
         \label{fig:normal_beat_samples}
     \end{subfigure}
     \hfill
     \begin{subfigure}[b]{0.45\linewidth}
         \centering
         \includegraphics[width=\textwidth]{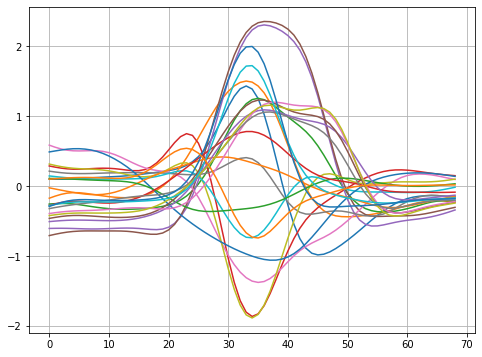}
         \caption{Ectopic QRS Complexes}
         \label{fig:ectopic_beats_samples}
     \end{subfigure}
     \caption{Examples from the Learning Set}
     \label{fig:learning_samples}
\end{figure}

\noindent
For training and testing purposes, we utilized the VPNet dataset \citep{Kovacs_data}, which consists of QRS complexes. This balanced subset of the MIT-BIH database \citep{MIT_database} includes only healthy and ectopic beats. The training-testing split adheres to the protocol defined by \cite{subsample_features}, ensuring that samples from the same patients are not present in both the training and testing sets. This separation guarantees that the results generalize well to new samples. The training set comprises 8,520 samples, while the test set contains 6,440 samples. Examples from the dataset are illustrated in Figure~\ref{fig:learning_samples}.

For this study, the VPNet dataset was further divided into three parts: $40\%$ of the training set was allocated for training, and the remaining $60\%$ was designated as a validation set, primarily used to tune the hyperparameters of various classifiers. The original test set remained unchanged and was solely used to evaluate classification accuracy.

\begin{sloppypar}
The samples in the dataset were processed using standard procedures \citep{ECG_data_analysis_book1,ECG_data_analysis_book2,ECG_data_analysis_paper2}. Initially, low-pass and high-pass filters with cutoff frequencies of $20$ Hz and $0.5$ Hz, respectively, were applied to remove noise and baseline shifts. Subsequently, the signals were standardized to have a zero mean and normalized by their maximum value. The QRS peaks were identified, and each sample was segmented to include 30 data points centered on the QRS complex peak, corresponding to $L=30$ in Eq.~\eqref{eq:augmented_Y}.
\end{sloppypar}

\section{Experimental Results}
\label{sec:results}

\subsection{Transformation Example Using the TwoLeadECG Dataset}
\label{sec:illustr}

\noindent
Before presenting the classification results, we illustrate the transformation process of the LLT algorithm as implemented in the proposed LLT-ECG method. For this purpose, we utilize the TwoLeadECG dataset \citep{TwoLeadECG}, which comprises 523 training signals and 629 test signals, each consisting of 82 consecutive measurements.

\begin{figure}[H] \caption{Accuracy of each percentile pair in linear discrimination} \label{fig:2} \centering \includegraphics[width=0.5\textwidth]{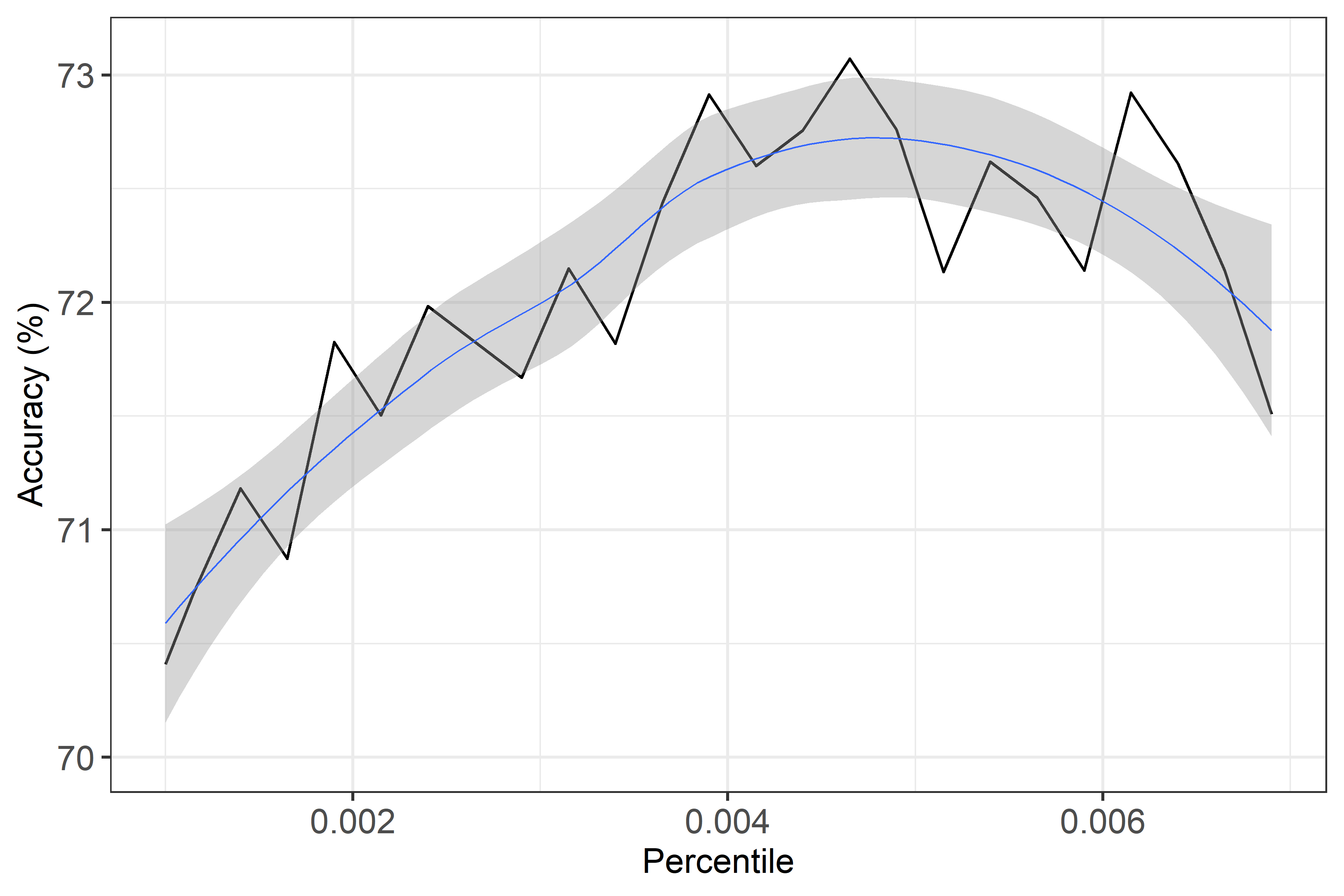} \end{figure}

We begin by defining the linear laws for the test set signals using an 11-length rolling time window.\footnote{This involved performing a 6th-order time-delay embedding of the 11-length series, resulting in a symmetric matrix where the last measurement occupies the final diagonal position.} This procedure yielded 72 laws per signal, amounting to a total of 45,288 laws across all test signals. These laws were stored in two matrices corresponding to the two signal classes. To increase the classification challenge and simulate streaming data, we randomly selected a single 11-length series from each test signal. We then took the first test signal, applied a 6th-order time-delay embedding, and computed its product with both law matrices. The results were squared to ensure all values were non-negative.\footnote{These calculations were performed using the LLT R package \citep{kurbucz2024llt}.} We computed the column means of the squared matrices and stored 25 percentiles ($p^o_c$, where $o \in {1, 2, \dots, 25}$) for each class $c$.\footnote{The column means represent the ``goodness of fit'' of the laws; lower percentiles correspond to better-fitting laws.} Finally, we evaluated the accuracy of each percentile pair as a linear separator between the two classes ($c \in {1, 2}$) using the following equation:

\begin{equation}
\begin{split}
 (Count( ~ p^o_1 ~ \leq~ p^o_2 * \alpha ~ | ~ c = 1 ~) ~ + \\
    Count( ~ p^o_1 ~ > ~ p^o_2 * \alpha ~ | ~ c = 2 ~)) ~ / ~ 523,
\label{eq:perc}
\end{split}
\end{equation}

\noindent
where $\alpha \in {0.01, 0.02, \dots, 2.00}$, and $N$ is the total number of signals (in this case, $N = 629$). The accuracy of each percentile pair is depicted in Figure \ref{fig:2}.

\begin{figure}[H] \caption{Discriminative power of the best percentile pair} \label{fig:3} \centering \includegraphics[width=0.5\textwidth]{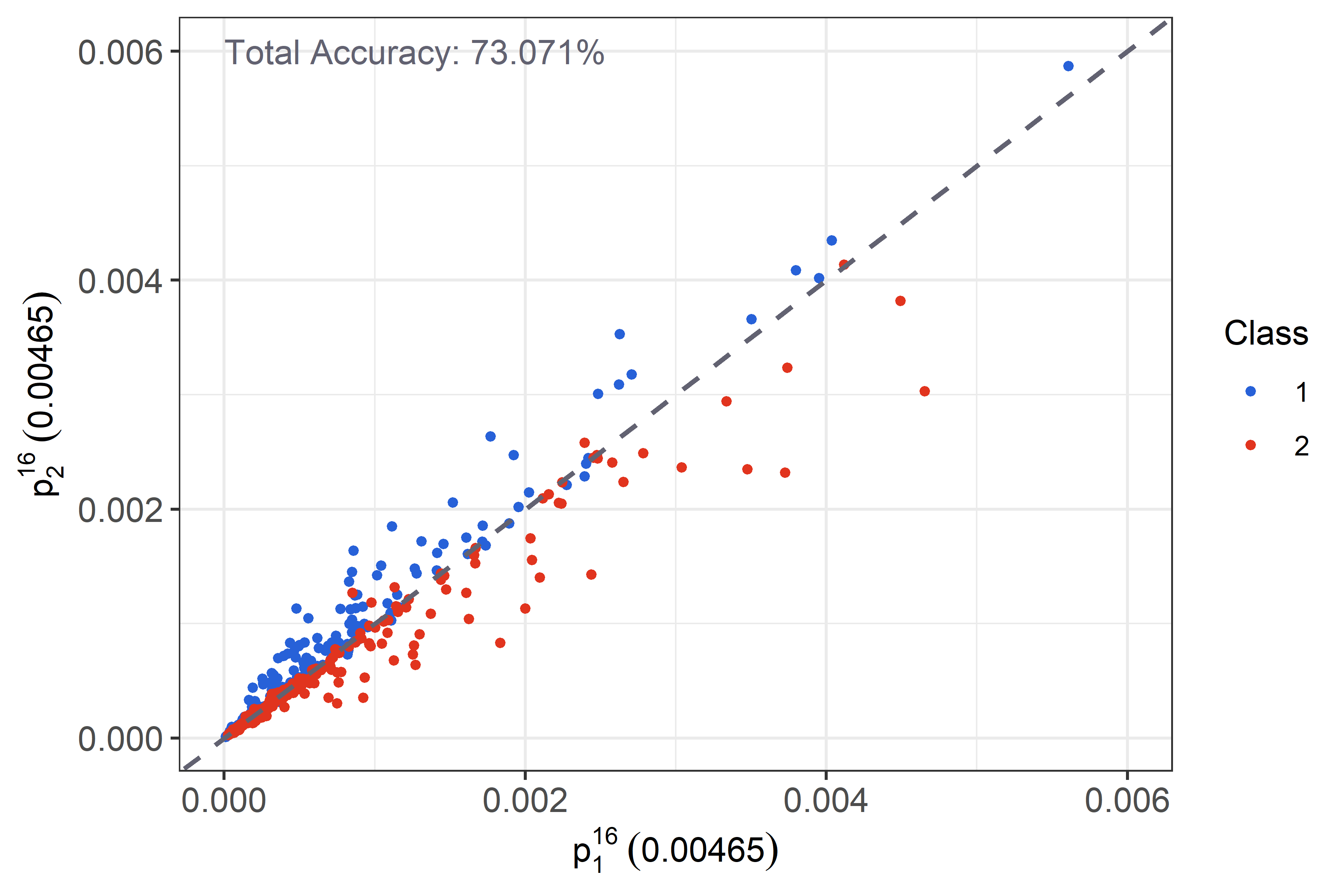} \end{figure}

As shown in Figure \ref{fig:2}, the 16th percentiles ($p^{16}_1$ and $p^{16}_2$) achieved the best separation, reaching a total accuracy of 73.07\% with $\alpha = 0.97$. The discriminative power of these two features is illustrated in Figure \ref{fig:3}.

For comparison, we trained 34 variations of 11 standard classifiers, both with and without dimensionality reduction techniques such as PCA, kernel PCA \citep{scholkopf1998nonlinear}, uniform manifold approximation and projection (UMAP) \citep{mcinnes2018umap}, and t-distributed stochastic neighbor embedding (t-SNE) \citep{van2008visualizing}, on the test set.\footnote{Each classifier was evaluated using 10-fold cross-validation. The KPCA, UMAP, and t-SNE transformations were performed using the \texttt{kernlab}, \texttt{umap}, and \texttt{Rtsne} R packages, respectively. t-SNE yielded 2 components, while the other methods provided 5 components.} Table \ref{tab:accu} summarizes their average and maximum accuracies.

\begin{table}[h!]
\centering
\caption{Test set accuracies of classifiers.}
\label{tab:accu}
\resizebox{0.825\textwidth}{!}{
\begin{tabular}{|l||c|c||c|c|}
\hline
\multicolumn{1}{|c||}{\multirow{2}{*}{Classifier}} & Num.    & \multicolumn{1}{|c||}{Versions}                        & Mean            & Max.         \\
\multicolumn{1}{|c||}{}                            & of ver. &   \multicolumn{1}{|c||}{(best is italics)}                                                        & acc. (\%)       & acc. (\%)       \\ \hline
Binary GLM Logistic Regression                    & 1       &                                                           & 49.452          & 49.452          \\ \hline
Binary GLM Logistic Regression (PCA)              & 1       &                                                           & 52.895          & 52.895          \\ \hline
Binary GLM Logistic Regression (KPCA)              & 1       &                                                           & 52.895          & 52.895          \\ \hline
Binary GLM Logistic Regression (UMAP)              & 1       &                                                           & 50.704          & 50.704          \\ \hline
Binary GLM Logistic Regression (t-SNE)              & 1       &                                                           & 55.399          & 55.399          \\ \hline
Discriminant                                      & 2       & linear, \textit{quadratic}                                         & 51.487          & 53.678          \\ \hline
Discriminant (PCA)                                & 2       & \textit{linear}, \textit{quadratic}                                         & 52.895          & 52.895          \\ \hline
Discriminant (KPCA)                                & 2       & linear, \textit{quadratic}                                         & 53.521          & 54.304          \\ \hline
Discriminant (UMAP)                                & 2       & linear, \textit{quadratic}                                         & 53.834          & 56.964          \\ \hline
Discriminant (t-SNE)                                & 2       & linear, \textit{quadratic}                                         & 55.086          & 55.399          \\ \hline
Efficient Linear SVM                              & 1       &                                                           & 48.357          & 48.357          \\ \hline
Efficient Linear SVM (PCA)                        & 1       &                                                           & 50.861          & 50.861          \\ \hline
Efficient Linear SVM (KPCA)                        & 1       &                                                           & 53.678          & 53.678          \\ \hline
Efficient Linear SVM (UMAP)                        & 1       &                                                           & 51.017          & 51.017          \\ \hline
Efficient Linear SVM (t-SNE)                        & 1       &                                                           & 53.678          & 53.678          \\ \hline
Efficient Logistic Regression                     & 1       &                                                           & 48.357          & 48.357          \\ \hline
Efficient Logistic Regression (PCA)               & 1       &                                                           & 52.895          & 52.895          \\ \hline
Efficient Logistic Regression (KPCA)               & 1       &                                                           & 53.365          & 53.365          \\ \hline
Efficient Logistic Regression (UMAP)               & 1       &                                                           & 50.704          & 50.704          \\ \hline
Efficient Logistic Regression (t-SNE)               & 1       &                                                           & 55.399         & 55.399          \\ \hline
Ensemble                                          & 5       & bagged-, boosted-, RUSboosted trees, \textit{subs. KNN}, -disc.    & 65.509          & \textbf{78.091} \\ \hline
Ensemble (PCA)                                    & 5       & \textit{bagged-}, boosted-, RUSboosted trees, subs. KNN, -disc.    & 59.906          & 67.919          \\ \hline
Ensemble (KPCA)                                    & 5       & \textit{bagged-}, boosted-, RUSboosted trees, subs. KNN, -disc.    & 62.754          & 68.075          \\ \hline
Ensemble (UMAP)                                    & 5       & \textit{bagged-}, boosted-, RUSboosted trees, subs. KNN, -disc.    & 61.565          & 70.110          \\ \hline
Ensemble (t-SNE)                                    & 5       & \textit{bagged-}, boosted-, RUSboosted trees, subs. KNN, -disc.    & 61.064          & 69.640          \\ \hline
Kernel                                            & 2       & logistic regression, \textit{SVM}                                  & 69.405          & 71.049          \\ \hline
Kernel (PCA)                                      & 2       & logistic regression, \textit{SVM}                                  & 62.363          & 62.441          \\ \hline
Kernel (KPCA)                                      & 2       & \textit{logistic regression}, SVM                                  & 62.676          & 63.850          \\ \hline
Kernel (UMAP)                                      & 2       & logistic regression, \textit{SVM}                                  & 65.102          & 65.101          \\ \hline
Kernel (t-SNE)                                      & 2       & \textit{logistic regression}, SVM                                  & 54.851          & 56.182          \\ \hline
KNN                                               & 6       & fine, medium, coarse, cosine, cubic, \textit{weighted}             & 69.014          & \textbf{78.404} \\ \hline
KNN (PCA)                                         & 6       & fine, medium, coarse, cosine, \textit{cubic}, weighted             & 63.589          & 69.484          \\ \hline
KNN (KPCA)                                         & 6       & fine, medium, coarse, cosine, cubic, \textit{weighted}             & 63.250          & 66.510          \\ \hline
KNN (UMAP)                                         & 6       & fine, medium, coarse, cosine, cubic, \textit{weighted}             & 63.485          & 67.762          \\ \hline
KNN (t-SNE)                                         & 6       & \textit{fine}, medium, coarse, cosine, cubic, weighted             & 65.023          & 71.987          \\ \hline
Naive Bayes                                       & 2       & Gaussian, \textit{kernel}                                          & 55.008          & 56.964          \\ \hline
Naive Bayes (PCA)                                 & 2       & Gaussian, \textit{kernel}                                         & 54.617          & 56.495          \\ \hline
Naive Bayes (KPCA)                                 & 2       & Gaussian, \textit{kernel}                                         & 55.947          & 58.842          \\ \hline
Naive Bayes (UMAP)                                 & 2       & Gaussian, \textit{kernel}                                         & 53.443          & 53.521          \\ \hline
Naive Bayes (t-SNE)                                 & 2       & \textit{Gaussian}, kernel                                         & 54.773          & 55.086          \\ \hline
NN                                    & 5       & narrow, medium, \textit{wide}, bilayered, trilayered               & \textbf{74.773} & \textbf{77.465} \\ \hline
NN (PCA)                              & 5       & narrow, medium, \textit{wide}, bilayered, trilayered               & 71.487          & \textbf{73.396} \\ \hline
NN (KPCA)                              & 5       & narrow, \textit{medium}, wide, bilayered, trilayered               & 71.362          & \textbf{74.961} \\ \hline
NN (UMAP)                              & 5       & narrow, medium, \textit{wide}, bilayered, trilayered               & 68.263          & 71.362 \\ \hline
NN (t-SNE)                              & 5       & narrow, medium, \textit{wide}, bilayered, trilayered               & 71.424          & \textbf{75.743} \\ \hline
SVM                                               & 6       & linear, quadratic, cubic, \textit{fine-}, medium-, coarse Gaussian & 58.764          & \textbf{79.030} \\ \hline
SVM (PCA)                                         & 6       & linear, quadratic, cubic, \textit{fine-}, medium-, coarse Gaussian & 53.365          & 65.102          \\ \hline
SVM (KPCA)                                         & 6       & linear, quadratic, cubic, \textit{fine-}, medium-, coarse Gaussian & 57.512          & 69.640          \\ \hline
SVM (UMAP)                                         & 6       & linear, quadratic, cubic, \textit{fine-}, medium-, coarse Gaussian & 53.469          & 60.720          \\ \hline
SVM (t-SNE)                                         & 6       & linear, quadratic, cubic, \textit{fine-}, medium-, coarse Gaussian & 53.599          & 65.571          \\ \hline
Tree                                              & 3       & \textit{fine}, medium, coarse                                      & 61.189          & 68.232          \\ \hline
Tree (PCA)                                        & 3       & \textit{fine}, medium, coarse                                      & 61.189          & 65.884          \\ \hline
Tree (KPCA)                                        & 3       & \textit{fine}, medium, coarse                                      & 61.242          & 67.293          \\ \hline
Tree (UMAP)                                        & 3       & \textit{fine}, medium, coarse                                      & 61.607          & 69.953          \\ \hline
Tree (t-SNE)                                        & 3       & \textit{fine}, medium, coarse                                      & 62.493          & 67.293          \\ \hline
\end{tabular}
}
\end{table}

\subsection{Classification Results on the VPNet Dataset}

\noindent
Here, we present the results of the LLT-ECG method compared to both simple learning methods—such as random forests (RF) \citep{RF}, k-nearest neighbors (KNN) \citep{KNN}, support vector machines (SVM) \citep{SVM}, and a basic neural network (NN) \citep{NN1,NN2, NN3}—as well as more complex classifiers, including deep neural networks (DNNs).

\subsubsection{Simple Learning Methods}

\noindent
The results of LLT-ECG-based classification of ECG signals are summarized in Table~\ref{tab:results}. Standard performance metrics were used to evaluate the different classifiers. Among these metrics, the total accuracy (ACC) is defined as: 

\begin{equation}
    \textbf{ACC} = \frac{\text{TP} + \text{TN} }{\text{TP} + \text{FN} + \text{TN} + \text{FP}},
    \label{eq:acc}
\end{equation}

\noindent
where TP, TN, FP, and FN denote the numbers of true positives, true negatives, false positives, and false negatives, respectively. Sensitivity (Se), also known as recall, is defined as:

\begin{equation}
    \textbf{Se} = \frac{\text{TP} }{\text{TP} + \text{FN}}. 
    \label{eq:se}
\end{equation}

\noindent 
Precision (+P), also referred to as positive predictive value, is defined as: 

\begin{equation}
    \textbf{+P} = \frac{\text{TP} }{\text{TP} + \text{FP}}. 
    \label{eq:pp}
\end{equation}

These performance metrics were calculated for both the validation and test sets to examine how well the LLT-generated features generalize to a new group of patients. The validation set contains ECG signals from the same patients as the training set, whereas the test set comprises signals from different patients than those in the training and validation sets. These results characterize the real-world performance of the proposed method on new, previously unseen data. As observed in Table~\ref{tab:results}, the metrics are similar for both sets, falling within the range of state-of-the-art methods \citep{survey}.

\begin{table}

\centering
\caption{\label{tab:results}  LLT-ECG-based classification performance of simple methods evaluated on the VPNet dataset~\citep{Kovacs_data}. The general performance ranges of state-of-the-art methods, which are not necessarily trained on the same dataset, are also presented \citep{survey}. The highest value in each column is highlighted.}
{\resizebox{\textwidth}{!}{
\begin{tabular}{|l||c|cc|cc||c|cc|cc|} 
\hline
\multicolumn{1}{|c||}{\multirow{3}{*}{Method}} & \multicolumn{5}{c||}{Validation}                                                                                                             & \multicolumn{5}{c|}{Test}                                                                                                        \\ 
\cline{2-11}
\multicolumn{1}{|c||}{}                        & \multirow{2}{*}{\begin{tabular}[c]{@{}c@{}}Total\\Accuracy\end{tabular}} & \multicolumn{2}{c|}{Normal} & \multicolumn{2}{c||}{Ectopic} & \multirow{2}{*}{\begin{tabular}[c]{@{}c@{}}Total\\Accuracy\end{tabular}} & \multicolumn{2}{c|}{Normal} & \multicolumn{2}{c|}{Ectopic}  \\
\multicolumn{1}{|c||}{}                        &                                                                          & Se     & +P                 & Se     & +P                   &                                                                          & Se      & +P                & Se      & +P                  \\ 
\hline
RF                                             & 93.6\%                                                                   & 94.3\% & 93.1\%             & 93.0\% & 94.2\%               & 92.1\%                                                                   & 92.9\%  & 91.4\%            & 91.2\%  & 92.8\%              \\
SVM                                            & 95.0\%                                                                   & 96.3\% & 93.8\%             & 93.6\% & 96.2\%               & \textbf{94.3\%}                                                                  & 94.4\%  & \textbf{94.2\%}            & \textbf{94.2\%}  & 94.4\%              \\
SVM (linear)                                   & 89.4\%                                                                   & 89.9\% & 89.0\%             & 88.9\% & 89.8\%               & 91.8\%                                                                   & 93.2\%  & 90.6\%            & 90.4\%  & 93.0\%              \\
NN                                             & 95.2\%                                                                   & 95.7\% & 94.7\%             & 94.7\% & 95.6\%               & 93.1\%                                                                   & 94.0\%  & 92.3\%            & 92.2\%  & 93.9\%              \\
KNN (k=4)                                      & \textbf{96.4\%}                                                                   & \textbf{97.4\%} & \textbf{95.5\%}             & \textbf{95.4\%} & \textbf{97.4\%}               & 91.5\%                                                                   & \textbf{95.0\%}  & 88.8\%            & 88.0\%  & \textbf{94.6\% }             \\
State-of-the-art                             & N/A                                                                      & N/A    & N/A                & N/A    & N/A                  & N/A                                                                      & 80-99\% & 85-99\%           & 77-96\% & 63-99\%             \\
\hline
\end{tabular}}}
\end{table}

\subsubsection{Deep Neural Networks}

\noindent
Table~\ref{tab:clas_res} summarizes the classification accuracy, specificity (Sp), and sensitivity (Se) for six methods: a CNN, a spiking neural network (SNN), the spiking CNN proposed in~\cite{cinc22}, VPNet~\citep{Kovacs_data}, and two DNNs. The first group of methods was trained on the balanced VPNet dataset, as detailed in Section~\ref{sec:vpnet_dataset}. In contrast, the state-of-the-art DNN architectures in the second group were trained on an imbalanced version of the VPNet dataset, where the majority class (normal) was not downsampled to match the number of ectopic samples. Both groups followed the same training-test split protocol as used in the balanced dataset version~\citep{subsample_features}, ensuring the results in Table~\ref{tab:results} and Table~\ref{tab:clas_res} are comparable.

\begin{table}[h]
\caption{\label{tab:clas_res} Classification performances of DNNs on the test set.}
\vspace{4 mm}
\centering
\resizebox{0.7\textwidth}{!}{
\centerline{\begin{tabular}{|l||c||c|c|c|} \hline
\multicolumn{1}{|c||}{Architecture}   & Number of parameters & Accuracy    & Sp  & Se  \\ \hline
CNN~\citep{cinc22} & $212610$ &$95.92\%$    &$96.09\%$  &$95.75\%$ \\
SNN~\citep{cinc22} & $58880$ &$95.59\%$    &$93.73\%$  &$\textbf{97.45}\%$ \\
SCNN~\citep{cinc22} & $376704$ & $95.42\%$    &$95.31\%$  &$95.53\%$ \\
VPNet~\citep{Kovacs_data} & $39$ & $96.65\%$   & $96.83\%$  &$96.61\%$ \\
\hline
raw data + DNN~\citep{jbhi_veb} & $61900$ & $99.70\%$ & $99.89\%$ & $\textbf{97.68\%}$\\
6 features + DNN~\citep{icmla} & $11700$ & $99.41\%$ & N/A & $96.08\%$\\ \hline
\end{tabular}}
}
\end{table}

\section{Discussion}
\label{sec:disc}

\subsection{Results on the TwoLeadECG Dataset}

\noindent
According to Table~\ref{tab:accu}, only a few algorithms---highlighted in bold for their superior accuracy---outperformed the proposed method, which achieved an accuracy of $73.071$\% using a simple linear discriminator. This success is attributed to the method's ability to transform the original dataset into a nearly linearly separable feature space (see Figure~\ref{fig:3}). Notably, the proposed method made classifications using only two features, while the benchmark models relied on 11 input features in the non-transformed case, 2 in t-SNE, and 5 in other transformation methods. These results highlight the potential of the LLT-ECG method, as it not only delivers an accurate solution for ECG classification but also applies the LLT method to perform a verifiable transformation. Given its high accuracy, even with simple and transparent classifiers, this method could serve as a foundation for new algorithms designed to support transparent and reliable decision-making.

\subsection{Results on the VPNet Dataset}

\noindent
In our second experiment (see Table~\ref{tab:results}), we first applied several simple classifiers within the LLT-ECG framework and compared their performance to state-of-the-art methods. We then extended the comparison to include more complex deep neural networks (DNNs) to evaluate the proposed method against advanced architectures.

\subsubsection{Simple Learning Methods}

\noindent
According to the results presented in Table~\ref{tab:results}, basic learning methods---such as RF, KNN, SVM, and NN---were effectively trained as part of the ECG-LLT framework. For the RF classifier, we balanced the number of estimators and tree depth to avoid overfitting. By selecting 10 estimators with trees of depth 6, we achieved comparable performance on both validation and test sets, indicating successful generalization.

The KNN classifier was optimized using the validation set, resulting in $k=4$ with the Chebyshev metric. This configuration yielded the highest overall accuracy on the validation set, although its performance on the test set was less convincing, particularly in terms of sensitivity for ectopic beats. Since KNN relies on proximity to training samples, new patient data in the test set that are farther from training samples in the feature space led to reduced accuracy. In contrast, NN and SVM can better generalize due to their ability to learn more complex patterns.

The SVM classifier, particularly with a nonlinear kernel, is well-suited to the LLT-transformed feature space, effectively finding separating hypersurfaces between clusters. It provided high and consistent accuracy, exceeding $94\%$ on both validation and test sets, demonstrating that the LLT-ECG successfully extracts meaningful features. Even a linear SVM achieved approximately $90\%$ accuracy, suggesting that the proposed method maps samples into distinct regions that are linearly separable.

Furthermore, a simple NN with one hidden layer---an input layer of 24 neurons, a hidden layer of 8 neurons, and an output layer of 2 neurons---was sufficient to achieve good classification performance. The NN generalized well, with similar accuracies on validation and test sets (Table~\ref{tab:results}). This simplicity indicates that the LLT-ECG effectively captures the necessary features.

\subsubsection{Deep Neural Networks}

\noindent
The proposed LLT-ECG classification method typically achieves a test accuracy that is about $\text{1-2}\%$ lower than the first group of methods and approximately $\text{5}\%$ lower than the DNN approaches. On the other hand, all of these methods require training via backpropagation, and most of them yield deep AI models with high complexity. In contrast, the proposed method operates without the need for error backpropagation, and the number of parameters, i.e., the length of the linear law ($n$), is equal to $11$. Moreover, as presented in Section~\ref{sec:illustr}, the LLT transformation extracts valuable features for classifying streaming data without the necessity for precise data alignment. This does not apply to its end-to-end DNN counterparts~\citep{jbhi_veb, icmla}, which segment and align heartbeats using the characteristic points of ECG waveforms (such as QRS, T, P). For this reason, improper peak detection and beat alignment may lead to high variability in the feature vectors~\citep{jbhi_veb}, potentially degrading classification accuracy on noisy data. The proposed LLT transformation, however, provides more robust features with decent discrimination power.

\subsection{Real-World Applicability and Limitations
\label{sec:limitations}}

\noindent
The primary computational complexity in constructing linear laws is determined by the column dimension of the matrix $Y_{ki}$ in Eq.~\eqref{eq:augmented_Y}. Specifically, the eigenvectors of the correlation matrix $C$ in Eq.~\eqref{eq:corrmat} can be computed using singular value decomposition (SVD), which requires $4mn^2 + 8n^2$ floating-point operations (flops) for an $m \times n$ matrix, where $m \geq n$ (see Table~8.6.1 in \cite{matcomp}). It is important to note that the column dimension is typically small and fixed prior to training, as it is equal to $11$ in our ECG case study. Consequently, linear laws can be efficiently updated with new data samples, unlike traditional DNNs that rely on backpropagation and necessitate retraining on the entire dataset. In practice, new time-embedded data samples only increase the row dimension of $Y_{ki}$ in Eq.~\eqref{eq:augmented_Y}, without impacting the column dimension that primarily governs computational complexity. This characteristic renders LLT features well-suited for incremental learning scenarios, such as clinical applications, where substantial amounts of physiological data are continuously collected.

However, the LLT-ECG-based approach presents certain limitations within the context of ECG applications. Although it has demonstrated robustness against various time series distortions, including noise and amplitude scaling \cite[see][]{LLT}, it remains sensitive to scaling along the time axis---a common occurrence in ECG signals \citep{keogh1997fast}. Another limitation pertains to the continuous filtration of laws obtained through streamed data classification. While employing a time-window approach can help maintain the method's speed and lightweight nature, ensuring an efficient algorithm that is free from catastrophic forgetting necessitates the inclusion of a law screening step following each classification.

Lastly, it is important to highlight that the proposed LLT features capture the similarity between the time embeddings of the input data and the extracted linear laws. In this regard, the LLT-ECG method embodies mathematical transparency, a form of interpretability \cite{Lipton}. Nevertheless, the clinical explainability concerning how the linear laws correlate with medical features remains an open question. To elucidate such relationships, post-hoc explanation methods can be utilized to evaluate the relevance between input features and the model's output after training. For example, similar to the approach in \cite{nicolai}, the average attribution per class could be computed for each linear law using the integrated gradients method \cite{IG}.

\section{Conclusions and Future Work}
\label{sec:conclusion}

\noindent
In this paper, we introduced a novel technique for learning features from time series data and successfully applied it to the task of binary ECG signal classification. This new method extracts features in a data-driven, forward manner, resulting in a classifier-agnostic feature space. These characteristics are achieved through the use of the principle of linear laws and the LLT method \citep{LLT, LLT_mech}. Linear laws are defined as common linear relationships among points in samples that belong to the same class. As a result, linear laws provide a concise and effective representation of classes. Moreover, the proposed LLT-ECG transformation offers a lightweight feature learning procedure that avoids the need for resource-intensive training via backpropagation, thereby eliminating the vanishing gradient problem. Additionally, the LLT-based approach demonstrated high data efficiency, achieving performance comparable to state-of-the-art methods while using less than half the training samples.

Future work on the LLT-ECG method could focus on two main research areas: developing techniques that build on the core ideas presented in this paper to enhance the medical interpretability of the classifications, and exploring self-selection mechanisms for the laws to ensure LLT-ECG remains up-to-date and lightweight, even in streaming data classification tasks. In addition to these research directions, we plan to release the LLT-ECG Python package to serve as a foundation for further development. Beyond ECG data, the original LLT algorithm has already shown success in various time series classification tasks, such as human activity recognition \citep{LLT} and price movement prediction \citep{kurbucz2023predicting}. In our future research, we plan to explore additional applications, including the classification of visually evoked potentials \citep{VEP} and tire sensor data \citep{roadsurface}. Since these signals exhibit morphological characteristics similar to ECG data, we expect that our method will perform well in the corresponding classification and regression tasks.

\section*{Data Availability}

\noindent
All the raw data generated in this study are available from the corresponding author upon reasonable request.

\section*{Acknowledgements}

\noindent
Project no. PD142593 was implemented with the support provided by the Ministry of Culture and Innovation of Hungary from the National Research, Development, and Innovation Fund, financed under the PD\_22 ``OTKA'' funding scheme. P.K. was supported by the \'UNKP-22-5 New National Excellence Program of the Ministry for Culture and Innovation from the source of the National Research, Development and Innovation Fund. Project no. TKP2021-NVA-29 and K146721 have been implemented with the support provided by the Ministry of Culture and Innovation of Hungary from the National Research, Development and Innovation Fund, financed under the TKP2021-NVA and the K\_23 `OTKA'' funding schemes, respectively.

\section*{Author Contributions Statement}

\noindent
P.P., M.T.K., P.K., and A.J. conceptualized the work and contributed to the writing and editing of the manuscript, as well as to the analysis. P.P., M.T.K., and P.K. acquired the data, and A.J. supervised the research.

\section*{Competing Interests}

\noindent
The authors declare no competing interests.

\bibliography{sample}
\end{document}